\documentclass[ERTS, 2026]{ERTS} 
\usepackage[T1]{fontenc}
\usepackage{graphicx}
\usepackage{amsmath}

\makeatletter
\AddToHook{cmd/maketitle/before}{\let\thetitle\@title}
\makeatother

\usepackage{float}

\usepackage{makecell}
\usepackage{multirow}
\usepackage{graphicx} 
\usepackage[usenames,dvipsnames]{xcolor}
\usepackage[colorlinks=true,linkcolor=blue,urlcolor=NavyBlue,citecolor=Fuchsia]{hyperref}
\usepackage{amsmath, amssymb}   
\usepackage{graphics}
\usepackage{wrapfig}
\usepackage{caption}
\usepackage{verbatim}
\usepackage{paralist}
\usepackage{cancel}
\usepackage{xspace}
\usepackage{xcolor}
\usepackage{multicol}
\usepackage{subcaption,url}
\usepackage{algorithm}
\usepackage{algpseudocode}
\usepackage{tabularx}
\usepackage{diagbox}
\usepackage{booktabs}

\usepackage{tikz,ifthen,pgfplots}
\usepackage[dvipsnames]{xcolor}
\usepackage{listings}

\newcommand\YAMLcolonstyle{\color{red}\mdseries}
\newcommand\YAMLkeystyle{\color{black}\bfseries}
\newcommand\YAMLvaluestyle{\color{blue}\mdseries}

\makeatletter

\newcommand\language@yaml{yaml}

\expandafter\expandafter\expandafter\lstdefinelanguage
\expandafter{\language@yaml}
{
  keywords={true,false,null,y,n},
  keywordstyle=\color{darkgray}\bfseries,
  basicstyle=\YAMLkeystyle,                                 
  sensitive=false,
  comment=[l]{\#},
  morecomment=[s]{/*}{*/},
  commentstyle=\color{purple}\ttfamily,
  stringstyle=\YAMLvaluestyle\ttfamily,
  moredelim=[l][\color{orange}]{\&},
  moredelim=[l][\color{magenta}]{*},
  moredelim=**[il][\YAMLcolonstyle{:}\YAMLvaluestyle]{:},   
  morestring=[b]',
  morestring=[b]",
  literate =    {---}{{\ProcessThreeDashes}}3
                {>}{{\textcolor{red}\textgreater}}1     
                {|}{{\textcolor{red}\textbar}}1 
                {\ -\ }{{\mdseries\ -\ }}3,
}

\lst@AddToHook{EveryLine}{\ifx\lst@language\language@yaml\YAMLkeystyle\fi}
\makeatother

\newcommand\ProcessThreeDashes{\llap{\color{cyan}\mdseries-{-}-}}
\usetikzlibrary{arrows,trees,backgrounds,automata,shapes,decorations,plotmarks,fit,calc,positioning,shadows,chains}
\tikzstyle{every pin edge}=[<-,shorten <=1pt]
\tikzstyle{neuron}=[circle,fill=black!25,minimum size=17pt,inner sep=0pt]
\tikzstyle{input neuron}=[neuron, fill=green!40]
\tikzstyle{output neuron}=[neuron, fill=red!40]
\tikzstyle{hidden neuron}=[neuron, fill=blue!40]
\tikzstyle{constructed neuron}=[neuron, fill=orange!50]
\tikzstyle{annot} = [text width=6em, text centered]
\tikzstyle{nnedge} = [-{stealth},shorten >=0.1cm, shorten <=0.05cm,line width=0.8pt,black]
\usetikzlibrary{calc}

\makeatletter
\newcommand\footnoteref[1]{\protected@xdef\@thefnmark{\ref{#1}}\@footnotemark}
\makeatother

\newcommand{\lard}{\textsc{LARD}\xspace}
\newcommand{\lardun}{\textsc{LARD v1}\xspace}
\newcommand{\larddeux}{\textsc{LARD v2}\xspace}
\newcommand{\xplane}{X-Plane\xspace}

\newcommand{\arcgis}{ArcGIS\xspace}

\usepackage[colorinlistoftodos]{todonotes}

\begin{document}

\title{\lard 2.0: Enhanced Dataset and Benchmarking\\ for Autonomous Landing Systems
}

\author{%
    Yassine Bougacha\authorNumber{1},
    Geoffrey Delhomme\authorNumber{5},
    Mélanie Ducoffe\authorNumber{1}\authorNumber{2},
    Augustin Fuchs\authorNumber{3},
    Jean-Brice Ginestet\authorNumber{4},
    Jacques Girard\authorNumber{1,5},\\
    Sofiane Kra\"iem\authorNumber{3},
    Franck Mamalet\authorNumber{1},
    Vincent Mussot\authorNumber{1},
    Claire Pagetti\authorNumber{3},
    and Thierry Sammour\authorNumber{2}
\vspace{3pt}
}

\address{
$^1$ IRT Saint Exupéry, Toulouse, France -- firstname.lastname@irt-saintexupery.com,\\
$^2$ Airbus, Toulouse, France -- firstname.lastname@airbus.com,\\
$^3$ ONERA, Toulouse, France -- firstname.lastname@onera.fr,\\
$^4$ DGA, Toulouse, France -- firstname.lastname@intradef.gouv.fr,\\
$^5$ Thales, Toulouse, France -- firstname.lastname@thalesgroup.com
}

\maketitle

\chead{\lard 2.0: Enhanced Dataset and Benchmarking for Autonomous Landing Systems}

\pagestyle{fancy}

\thispagestyle{plain}
\licenseFootnote{Yassine Bougacha~et al}

\begin{abstract}%
  This paper addresses key challenges in the development of autonomous landing systems, focusing on dataset limitations for supervised training of Machine Learning (ML) models for object detection.
  Our main contributions include: (1) Enhancing dataset diversity, by advocating for the inclusion of new sources such as BingMap aerial images and Flight Simulator, to widen the generation scope of an existing dataset generator used to produce the \lard dataset; (2) Refining the Operational Design Domain (ODD), addressing issues like unrealistic landing scenarios and expanding coverage to multi-runway airports; (3) Benchmarking ML models for autonomous landing systems, introducing a framework for evaluating object detection subtask in a complex multi-instances setting, and providing associated open-source models as a baseline for AI models' performance.
\end{abstract}
\color{black}
\section{Introduction}
\label{sec:intro}

As interest in autonomous systems grows, 
machine learning (ML) vision-based algorithms are increasingly used.
One of the major challenges is the collection of sufficient and representative real-world data. In the field of autonomous landing systems in aerospace, despite significant practical and commercial interest, there remains a notable lack of open-source datasets containing real-world aerial images.
This shortage has led previous efforts to rely heavily on synthetic datasets. We can cite BARS in \xplane, \lard in Google Earth~\cite{chen2023bars,ducoffe2023lard}.
Notably, the \lard dataset, which provides high-quality aerial images collected from Google Earth for runway detection during the approach and landing phases, also offers manually labelled images from real landing footage. Although such a data set has contributed significantly to autonomous aerial tasks, several limitations persist.

One specificity of \lard compared to other dataset is the definition of a proper Operation Design Domain (ODD).
\lard introduced a useful framework for defining the operational domain coverage for autonomous systems, which has supported many AI certification efforts~\cite{shuaia2023advances,cappi2024design,denney2023assurance, luettig2024ml}. However, some aspects of this domain, particularly the sampling of images (such as the cone trajectory), lead to unrealistic landing scenarios. These scenarios make it difficult to assess the true performance of AI models in real operational domain, as they do not reflect real-world conditions. Furthermore, the initial version of \lard focused solely on single-runway airports, which is a significant limitation since 80\% of commercial air traffic is managed by airports with two or more runways.
The ODD for multi-runway detection has not been clearly defined in the literature.

Second, as highlighted in~\cite{de2024validating}, relying solely on simulated data is inadequate for training and validating navigation systems in complex airport environments. One key challenge lies in the lack of diversity within these datasets. A first step toward a solution would be to enhance the existing dataset by incorporating data from a variety of simulators. 

While there exist some open-source models available, like~\cite{gde-yolo-v8}, most prior vision-based landing papers do not release their model weights or code. A contributing factor to this gap is the lack of a clear and standardized definition of the ML tasks and adequate metrics for runway detection.
 While \lard, or BARS mentioned object detection as a key task, little exploration has been done to compare runway detection to general object detection frameworks and the efficiency of current models. For example,~\cite{li2024yolo} trained object detection models, but as they noted, detecting a single object (such as one runway) is less complex than multiple ones. Notably, they did not disclose their models. 

In this work, we will address the limitations identified above.
Subsequently, we will refer to as \lardun when discussing the~\cite{ducoffe2023lard} paper and dataset; and to \larddeux when presenting the contribution and enhanced dataset of this paper.
Our main contributions are the following:
\begin{itemize}
    \item \textbf{Refinement of the Operational Design Domain (ODD)}: We refine the definition of our previous ODD for vision-based landing by focusing on the approach segment, tightening and segmenting the approach cone, and extending it to multi-runway airports. We also introduce an \emph{Extended ODD} band with tolerance margins, which explicitly captures near-boundary poses around the nominal cone. 
    \item \textbf{Enhancing Dataset Diversity}: Recognizing the limitations of \lardun, we propose a second version available at~\cite{deelai-lard-v2}, built from a calibrated scenario generator that produces consistent images across multiple virtual globes (Google Earth, Bing Maps, \arcgis) and flight simulators (\xplane and Flight Simulator). The resulting dataset covers hundreds of major multi-runway airports with precise per-runway annotations and In ODD / Extended ODD labels.

    \item \textbf{ODD-aware benchmarking ML models for runway detection}: We propose an extended detection metric (\textit{e-mAP}) that accounts for both \emph{In and Extended ODD} runways, and use it to benchmark modern object detectors on this second version of \lard, publicly available at~\cite{deelai-lard-v2-models}. In particular, we compare training strategies that include or exclude \emph{Extended ODD} runways, we quantify the impact of simulator diversity using single-source, leave-one-out and full multi-source training configurations, and we  provide baseline results that can serve as reference models for future work revolving around runway detection.

\end{itemize}

The simulator and the dataset are available on github at~\cite{deelai-lard-v2}, while the models presented in this paper are provided in a separate repository~\cite{deelai-lard-v2-models}.


\section{Operational Design Domain}\label{sec:odd}
The Operational Design Domain (ODD)~\cite{j3016} is a description of the set of constraints under which a system is designed to operate. In our case, it includes the geometry of the landing or the type of airports and runways considered, but also factors that can affect optical sensors, and by extension ML component capabilities, such as time of day and weather conditions.

\subsection{Approach cone} 
In our work, we only consider the \emph{approach} segment 
which ranges from -6000~m to -280~m along-track distance from Landing Threshold Point (LTP). Throughout this segment, the acceptable aircraft positions remain similar but not equivalent to the ODD of \lardun~\cite{ducoffe2023lard}: the \emph{lateral path angle} lies within [-3°,3°], originating from the LTP, while the \emph{vertical path angle} takes its values in [-1.8°,-5.2°] w.r.t. the Vertical Reference Point (VRP), a point on the centerline located 305~m  beyond the LTP.

For the aircraft attitude, the range for the \emph{pitch}\footnote{\label{foot:acft_body_ref}Aircraft body reference} stays constant, between [-15°,~+5°], which translates to an average nose-down orientation of about 5° relative to the horizon. However, to define appropriate ranges for the aircraft \emph{roll}\footnoteref{foot:acft_body_ref} and \emph{yaw}\footnote{Runway true heading reference}, the approach cone is split into 3 segments as described in table~\ref{tab:yaw_roll_ranges}.

\begin{table}[ht]
    \centering
    \begin{tabular}{ccc}
        \toprule
        Along-track distance (m) & Yaw range (°) & Roll range (°) \\
        \midrule
        $[-6000,\,-4500]$ & $[-24,\;24]$   & $[-30,\;30]$ \\
        $[-4500,\,-2500]$ & $[-24,\;24]$   & $[-15,\;15]$ \\
        $[-2500,\,-280]$  & $[-18.5,\;18.5]$ & $[-10,\;10]$ \\
        \bottomrule
    \end{tabular}
    \caption{Allowed \emph{yaw} and \emph{roll} ranges as a function of along-track distance from the LTP}\label{tab:yaw_roll_ranges}
\end{table}


\subsection{Other hypotheses}
\begin{figure}[h!tb]
    \centering
    \includegraphics[width=.8\linewidth]{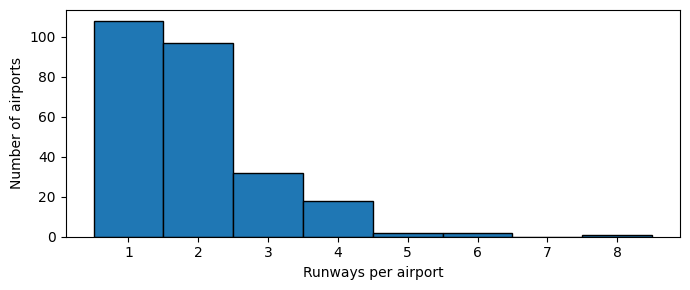}
    \caption{Distribution of runways per airports in \larddeux}
    \label{fig:distrib_per_airport}
\end{figure}

Unlike \lardun which handled only single-runway airports, the second version targets the world's major commercial ones:  First, we identified the 300 airports with the most traffic worldwide, then reduced this list according to the quality of the runway imagery we could obtain. Airports whose satellite images were badly pixelated or too poor for reliable analysis were filtered out, leaving us with 260 single- and multi-runway airports, distributed as shown in Figure~\ref{fig:distrib_per_airport}.
Thus, because most of these have multiple runways, we extended \lard to this more complex context, and had to define a rigorous approach for the runways labelling, described in Section~\ref{subsec:labelling}.

We preserve other ODD restrictions from \lardun, such as the presence of a \emph{piano} on the runway, and the optimal weather conditions. Despite using only the major commercial airports, the former constraint was still needed to remove smaller runways with different types of markings that may be present in these multi-runways airports. The latter constraint was preserved for easier comparison between our various sources of data, considering that some of these cannot simulate night images or adverse weather conditions. However, the addition of new types of simulators (especially \emph{\xplane} and \emph{Flight Simulator}) opens the door to the creation of dedicated dataset, allowing to test AI algorithms in more complex and realistic conditions.

\subsection{Intended function}
\begin{figure}[h!bt]
    \centering
    \includegraphics[width=\linewidth]{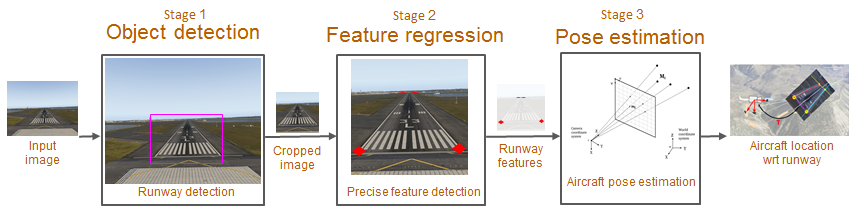}
    \caption{VBL constituent architecture with 3 stages}
    \label{fig:archi}
\end{figure}

As described in~\cite{ducoffe2023lard}, the ML constituent architecture should fulfil the intended function up to the pose estimation. 
We rely on a 3-stage architecture  directly inspired from 
Daedalaen AG~\cite{balduzzi2021neural} 
and illustrated in Figure~\ref{fig:archi}. 

\begin{figure*}[h!bt]
    \centering
    \includegraphics[width=.9\linewidth]{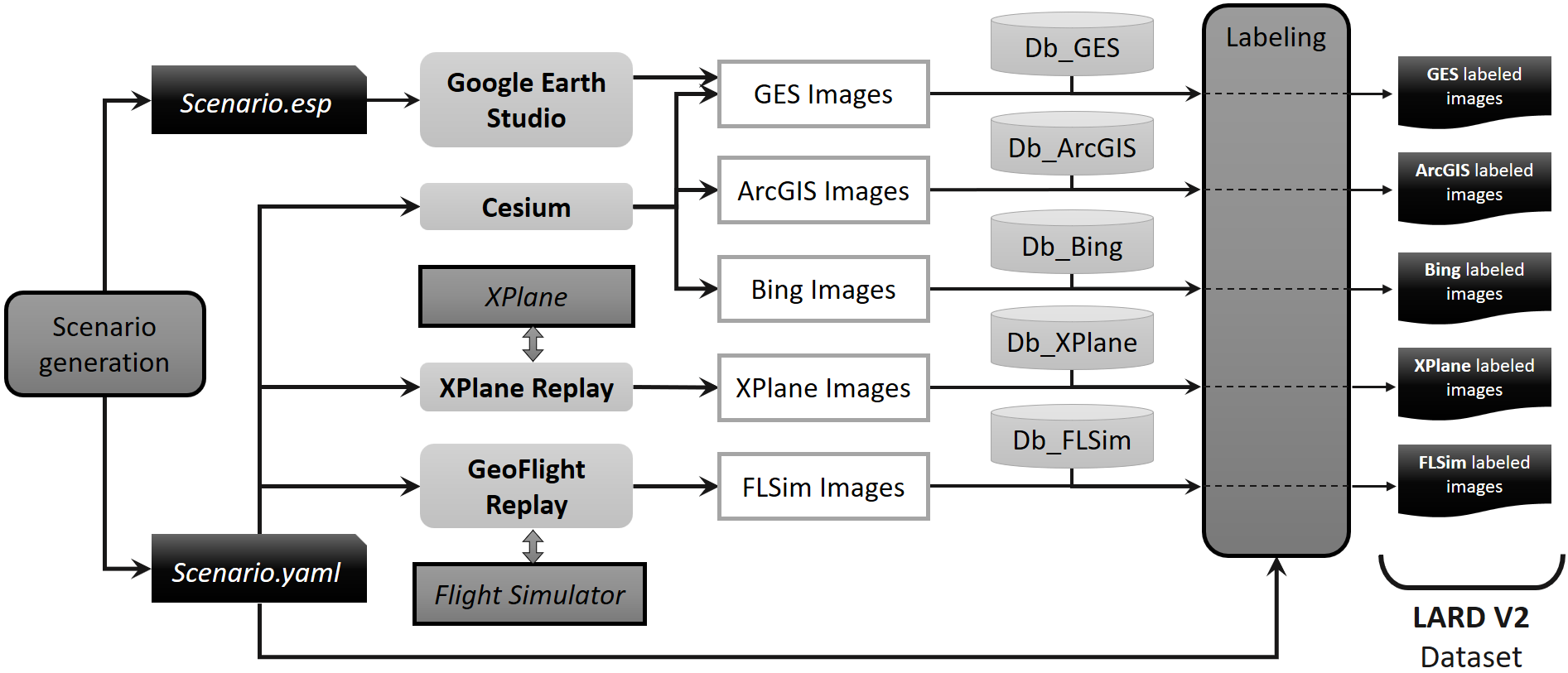}
    \caption{Lard V2 scenario generation and labeling workflow.}
    \label{fig:lard_V2_workflow}
\end{figure*}

The first stage is based on an object detection step that is in charge of computing a bounding box around the detected runway. The image is then cropped around the bounding box and a second stage is in charge of identifying key features on the runways, typically the 4 corners and specific runway contour lines. These keypoints then serve for the pose estimation stage, which can be done with a non-ML approach by the last stage.
As~\cite{ducoffe2023lard}, we focus only on the first object detection stage. 


\section{Synthetic data generators}

\lardun relies solely on Google Earth Studio to generate images. 
This tool, based on a 3D projection of Google Earth satellite imagery, produces images that may be partially deformed, or with various level of quality depending on the satellite coverage of the targeted area. This can be compensated by using multiple virtual globes and satellite sources for the same images. For this reason, 
we extended \lard to interface with a tool based on Cesium, which allows to generate images from different satellite layers, including Google Earth, \arcgis and Bing Maps. This interface also eliminates the need to go through the tedious process of generating images  through Google Earth Studio interface which is not initially designed for that purpose.

However, virtual globes alone are unable to simulate clouds and variable weather conditions, and they provide limited support for night scenes. Recent flight simulators, on the other hand, offer almost photorealistic environments with controllable atmospheric conditions and lighting. For this reason, we further extended \larddeux to interface with two such simulators, \xplane and Flight Simulator. Using the unified interface presented in Section~\ref{subsec:interop} allows us to generate, for each scenario, a family of images that includes heterogeneous satellite textures and more realistic simulator renderings.



\subsection{Interoperability}\label{subsec:interop}

Figure~\ref{fig:lard_V2_workflow} presents the overall workflow used to generate the \larddeux dataset and illustrates the interoperability of our tooling with different data sources.
Interoperability between the different tools is ensured by a common scenario description stored in a \emph{.yaml} file, which contains:
\begin{itemize}
    \item A header with global information, including the expected image size and field of view,
    \item A series of poses, defined by the position and attitude of the aircraft, the targeted runway and airport, and a date and time.
\end{itemize}

\begin{figure}[h!tb]
  \centering
  \begin{lstlisting}[language=yaml,basicstyle=\ttfamily\footnotesize,
frame=single]
airports_runways:
  LFBO:
    - 32R
    - 14L
    - 32L
    - 14R

image:
  height: 1024
  width: 1024
  fov_x: 60.0
  fov_y: 60.0

poses:
  - uuid: 468b7855-064c-473d-b0bd-b7bee9b26bab
    airport: LFBO
    runway: 14R
    pose: 
    - 1.3271, 
    - 43.6604, 
    - 286.1787, 
    - 140.4716, 
    - 86.1030, 
    - 6.7669
    time:
      second: 1
      minute: 0
      hour: 10
      day: 1
      month: 6
      year: 2020

runways_database: ./data/runways_db_V2_GES.json
trajectory:
  sample_number: 1
  \end{lstlisting}
  \caption{Example of \texttt{.yaml} scenario file used in \larddeux, showing an aircraft pose at  Toulouse Blagnac airport (LFBO), during an approach on runway 14R.}
  \label{lst:yaml_example}
\end{figure}

Figure~\ref{lst:yaml_example} shows a typical \texttt{.yaml} scenario used in \larddeux. The header encodes global information such as the list of airports and runways involved and the image geometry (resolution and field of view), while each element of the \texttt{poses} array specifies one camera pose with its associated airport, runway, six-degree-of-freedom position and attitude, and timestamp.

This scenario file is the output of the first generation step implemented in \lard. While the legacy tool Google Earth Studio used in \lardun still requires a specific file format\footnote{the \emph{.esp} file which can be generated alongside the \emph{.yaml} file.}, the \emph{.yaml} scenario which is directly consumed by all the new tools to render images in \larddeux, typically \xplane, Flight Simulator and virtual globe accessible through Cesium. 
A limitation of this approach is that the same geographical coordinates are not always mapped to exactly the same apparent positions in each tool, which can result in slight spatial offsets of objects in their respective 3D spaces. These discrepancies arise from several factors: imperfect mapping of satellite imagery onto the Earth, difference in elevation models and ground altitudes between providers, and geometric simplifications introduced in flight simulators. 
In our case, even a small misalignment of a few meters at a runway corner will make the annotation unusable, which is why we had to define the calibration procedure presented in section~\ref{subsec:calibration}.

In practice, the scenario generator provided in \larddeux allows to generate discrete poses along an approach trajectory, with the possibility to manually tune several aspects of this trajectory: each parameter (typically distance, lateral and vertical deviations and attitude constraints) can be sampled either from a bounded uniform distribution or from a normal distribution specified by a mean value and a standard deviation. For the construction of the \larddeux dataset, we opted for simple uniform draws within the ranges of the parameters defined in our ODD presented in Section~\ref{sec:odd}, with ten poses per ODD segment and per runway. 
Alternatively, 
users can bypass the generator and provide their own scenarios directly as lists of poses in the form of simple sequences of 6 degree of freedom positions. External tools that provide trajectories, for example from ADS-B recordings, can thus be converted in the same \emph{.yaml} representation and then rendered by our generators.

\subsection{Calibration}\label{subsec:calibration}

The starting point of our work is an open database~\cite{site_ourairports} containing the corners of the runways of all airports in the world. This database is highly valuable but inaccurate. In general, the latitude, longitude or altitude of the points provided do not match the exact corners of the runways. The position error usually represents a few meters but could be higher when the runways' markings are renewed due to maintenance, or repainting, or when the satellite imagery is misaligned. Moreover, even when the coordinates match a  specific data source, they will often be different for other sources, due to problems described in Section~\ref{subsec:interop}. This represents a major issue for our labels which must be as precise as possible, to ensure the feasibility of the various tasks and subtasks considered. 
This forced us to define a process to correct the coordinates of all the runways, resulting in a set of high-precision dedicated databases for each of the data sources. 


A natural first idea would be to rely directly on each rendering tool to obtain accurate runway coordinates. Unfortunately, this turns out to be impractical. In virtual globes, runways are only represented as textures on the terrain: the coordinates of runway corners are not exposed as structured objects and cannot be retrieved through queries. In flight simulators, one might expect internal navigation or scenery databases to provide this information, but in practice, these data are either inaccessible, only partially documented, or of insufficient quality. Moreover, such extraction strategy would not solve the problem for the virtual globes at all, for which no runway metadata is available. For all these reasons, we designed a unified  external calibration procedure. For a single data source (virtual globe or simulator), this procedure consists of four successive steps:
\begin{enumerate}
\item \textbf{Runway acquisition:} For each candidate threshold, we place a camera at a fixed altitude above the threshold center, facing the ground and aligned with the runway axis. We then capture the corresponding image using predefined field of view and image sizes, which guarantees that the \emph{piano} is fully visible near the center of the resulting picture. 
  \item \textbf{Data curation:} The resulting set of images can then be reviewed manually: frames with non-visible runways, or missing specific threshold markings can be discarded. This quality-control pass guarantees that the runways are kept only if their representation within this data source is accurate.
  \item \textbf{Automatic corner detection:} Our goal is to automatically find the corners at the bottom of the threshold (since the runway is facing up thanks to the camera orientation set at step 1). After a manual labelling of \emph{pianos} on 500 images, we were able to train an object detector (a Yolo-v11~\cite{yolo11_ultralytics}) on these markings. Because each image is centered on exactly one piano, the label obtained precisely contains the location of the two bottom corners of the runway.
    \item \textbf{Geometric projection:} Knowing the image metadata defined at step 1 (camera geodetic position, intrinsic parameters and pixel dimensions) and the pixel coordinates of the runway corners from step 3, we can compute their geodetic positions by back-projection, resulting in high-accuracy latitude and longitude for each runway end.
\end{enumerate}

It is worth noting that step 1 requires a way to precisely set the altitude to a specific value above ground level, to ensure the quality of the back-projection at step 4. It is usually feasible in the simulators and also using the Cesium-based tool, or using publicly available API like Google elevation API.

This process produces a specific database of runway coordinates that is tailored to each data source. As a result, we can generate images based on a single \emph{.yaml} file in multiple tools, and use the appropriate source-specific runways database during the labelling process to systematically obtain high quality labels.


\color{black}
\section{Dataset - \larddeux}
The second version of the LARD dataset is open source and available at~\cite{deelai-lard-v2}. This section describes its organisation and how automatic labelling is done to account for multiple runways.

\subsection{Strategy of dataset definition}

\larddeux is built from $1024 \times 1024$-pixel pictures captured in the three segments defined in Section~\ref{sec:odd}. This choice of size ensures that, at 6000~m, with a glide slope close to $3^\circ$ (ensured by the ODD), runways are still visible and recognisable. For each segment, we take 10 images at random positions and orientations within the acceptable ranges of the ODD, resulting in 30 pictures for every runway and generator. With close to 980 runways in the current list of airports, this corresponds, in principle, to roughly 30\,000 images per data source.

After completion of the generation and cleaning stages, however, \larddeux now contains around 110\,000 images across the five generators, ranging from around 18\,000 for Bing Maps, to 26\,000 for Flight Simulator. This is below the theoretical number of samples defined by our scenarios, because a non-negligible fraction of images had to be discarded during quality control. Typical rejection cases include poses located inside surrounding relief or below ground level in the elevation models, as well as sever artefacts in the virtual globes: heavily degraded or missing runway markings, and runway with implausible altitude profiles, exhibiting strong oscillations every few tens of meters that make both the runway boundaries and primary markings unusable for detection. Image generation for each back-end requires on the order of 15 to 40 hours on standard hardware, while, once the calibration procedure of Section~\ref{subsec:calibration} has been performed for a given source, the subsequent labelling of all images for that source is fully automatic and completes in a few minutes.

\begin{figure*}[h!tb]
  \centering
  \begin{tabular}{|c|c||c|}
    \hline
Input image RGB & $Img$ & shape (H,W,3)\\

Runway BB & $\mathcal{B}_i= (x_i,y_i,w_i,h_i)$ & $(x_i,y_i)$ center of the BB\\

Runway ODD indices (or flag) &  $o_i\in\{0,1\}$ & indicate if the runway is in the ODD \\

Predicted BB & $\hat{\mathcal{B}}_i= (\hat{x}_i,\hat{y}_i,\hat{w}_i,\hat{h}_i, \hat{s}_i)$ & $\hat{s}_i$ is confidence score \\

List of In ODD runways &  $I=\{i|o_i=1\}$ & \\
List of out-of-ODD runways &  $J=\{j|o_j=0\}$ & \\

List of predicted runways & $P=\{i|\hat{s}_i>0.5\}$ & can also use the NMS (Non Maximum Suppression) algorithm\\

    \hline
  \end{tabular}
    \caption{Notations to define mAP and extended \emph{mAP}  \label{tab:notations}}
\end{figure*}


Finally, it is worth noting that for model development, we decided to split the airport list in two halves, with approximately 50\% of airports for the training set\footnote{Dataset that will be used by the ML engineer to train and potentially validate their models before separate testing} and 50\% reserved for the test set, to prevent location-specific leakage, and ensure that runways seen during testing have not appeared during training.

\subsection{Labelling}\label{subsec:labelling}
A key challenge when dealing with multiple runways at an airport is to determine which runway to label in each image. To address this, we use the ODD defined in Section~\ref{sec:odd} as a reference. For each pose in a landing scenario, we consider all the runways that could appear in the camera's field of view based on the current position and orientation. Then we check if the aircraft is within the approach cone of each candidate runway, as specified in the ODD. The starting point for this is to determine the along-track distance between the aircraft and the targeted threshold, to identify the appropriate segment as defined in Table~\ref{tab:yaw_roll_ranges}. Once each parameter has been validated against each of the acceptable ranges, the runway can be labelled with an additional flag indicating that it is \emph{In ODD}.
Nevertheless, when looking into the details, this per-parameter range verification is less trivial than it appears. There are indeed some new challenges arising regarding the labelling process, originating from both the data source spatial offsets and the computation of orientation during generation:

\begin{itemize}
    \item \textbf{Data source offsets:} The scenarios are usually generated from a single data source, but the corresponding images will be rendered on multiple sources, where runways may have slight position differences, resulting in changes in relative positions. If the initial position was close to the border of the approach cone, there is a chance that this same position will be outside of the cone for a different source, which would result in an \emph{Out of ODD} flag.
    \item \textbf{Coupled rotations:} When rotations are combined during scenario generation, the pitch and roll are measured relative to the aircraft body, while the yaw is relative to the runway heading. As a result, these three parameters may impact each other during successive single-axis rotations. For instance, applying a rotation on the roll, followed by the pitch of an aircraft already alters the yaw relative to the runway.
\end{itemize}

To address these issues, we introduce tolerance margins around the acceptable angles and ranges, to account for these potential variations. 
Moreover, we defined an in-between \emph{Extended ODD} category, when any parameter exceeds the normal ranges but remains within twice the acceptable limits. This means that only the two categories \emph{In ODD} and \emph{Extended ODD} are considered valid 
situations where the runway should be recognisable by the model, but it also opens the door to a relaxed evaluation of the model performance, reducing the penalties when a model fails to detect a runway \emph{Extended ODD}.





\section{ML tasks and metrics}

In this paper, we focus exclusively on the first stage of the runway detection pipeline: the object detection task. This stage aims to locate and generate bounding boxes around runways present in aerial or satellite images. The output of this detection phase is critical, as it directly feeds into subsequent processing steps such as corner detection and aircraft pose estimation, but these downstream tasks are beyond the scope of the present study.

We consider several objectives that differ by the subset of data used during training. 
In particular, we define two complementary configurations that will structure the experiments in Section~\ref{sec:experiments}. First, we compare models trained only on \textit{In ODD} runways, with models trained on both \textit{In ODD} and \textit{Extended ODD} runways, in order to assess the impact of including near-boundary examples in the learning process. Second, we vary the composition of the training dataset with respect to the image generators, using models trained on a single data source, on all sources but one (\textit{leave-one-out}), or on the full combination, to quantify the influence of simulator diversity on generalization. 

\subsection{ML tasks}
Unlike the approach presented in the \lardun paper, which primarily addressed scenarios containing a single runway per image, our work extends this formulation to more complex scenes where multiple runways may be simultaneously present within the field of view. This reflects more realistic operational environments and introduces additional challenges for the detection models, such as differentiating between multiple adjacent or overlapping runways.

\subsection{Metrics - extended mAP}
\label{sec:emap}
To evaluate the performance of the object detection models in this extended context, we adopt the classical Mean Average Precision (mAP) metric, widely used in object detection benchmarks. However, in our case, the \emph{mAP} computation is adapted to handle extended runway scenarios, including runways that are not necessarily part of the Operational Design Domain (ODD). This ensures that the metric remains meaningful even when detecting runways that may not be relevant for the specific mission or operational constraints.
Figure~\ref{tab:notations} introduces some notations required for the metric.

The \emph{mAP} metric~\cite{schutze2008introduction,everingham2010pascal} is commonly used in object detection benchmarks (Note that since we have a single class 'runway', the \emph{mAP} coincide with the \emph{AP}). 
The \emph{mAP} metric evaluates the agreement between the predicted bounding boxes $\hat{\mathcal{B}}_i$ and the ground truth bounding boxes $\mathcal{B}_i$. Since the object detector is trained to detect runways that are within the ODD, the reference metric is computed using only the ODD runways:
$$
mAP_{\tau}\big((\mathcal{B}_i)_{i\in I},(\hat{\mathcal{B}}_p)_{p\in P}\big)
$$
However, since some runways outside the ODD (\emph{Extended ODD}) may also be visible and thus detected by the object detector, we additionally evaluate an extended version of the \emph{mAP} metric, which takes into account any possible detection of \emph{Extended ODD} runways:

$$
e\text{-}mAP_{\tau} =  \underset{J\subset\{j|o_j=0\}}{\max} mAP_{\tau}\big((\mathcal{B}_i)_{i\in I\cup J},(\hat{\mathcal{B}}_p)_{p\in P}\big)
$$ 

This extended metric \textit{e-mAP$_\tau$} searches, among all subsets $J$ of runways outside the nominal ODD, the subset that yields the highest possible score when these additional runways are treated as valid targets alongside the \textit{In ODD} ones (from the set $I$). It can bee seen as an oracle that, a posteriori, selects which \textit{Extended ODD} runways should be considered relevant for the evaluation, and reports the best $mAP$ value attainable for this selection. By construction, \textit{e-mAP$_{\tau}$} $\geq$ \textit{mAP$_{\tau}$} (the case where $J=\emptyset$ results in the standard \textit{mAP} on \textit{In ODD runways only}). 
This construction reduces the sensitivity of the score to the models' prediction on \textit{Extended ODD} runways, which are no longer automatically counted as hard false positives, but can instead be matched to a subset of extended ground truth runways if they are geometrically consistent. The \textit{e-mAP} thus represents a more stable basis for comparing models that differ in their behaviour near the ODD boundary. 



\section{Experiments}\label{sec:experiments}




\begin{table*}[ht]
\centering
\footnotesize
\begin{tabular}{l l c c l r r r r r r}
\toprule
\makecell{\textbf{Model: archi}\\\textbf{(Trained on)}} & \textbf{Score} &\textbf{Test GT}&   \textbf{GT} & \textbf{Detections} & \textbf{mAP} & \textbf{mAP50} & \textbf{mAP75} \\
\midrule
\multirow{5}{*}{\makecell{$M_{IN}$: yoloV8 \\\\ (IN\_ODD)}} & mAP & IN\_ODD  & 53610 & 80002 &  0.6974 & 0.9671 & 0.8122 \\
\cmidrule(lr){2-8}
& mAP & \makecell{IN\_ODD + \\ EXTENDED\_ODD} &  61788 & 80002 &  0.6498 & 0.9084 & 0.7539 \\
\cmidrule(lr){2-8}

& e-mAP & \makecell{IN\_ODD + \\ EXTENDED\_ODD} &  58242 & 80002 &  0.6882 & 0.9654 & 0.7999 \\
\midrule
\multirow{5}{*}{\makecell{$M_{EXT}$: yoloV8 \\\\ (IN\_ODD+ \\ EXTENDED\_ODD)}}  & mAP & IN\_ODD & 53610 & 100504 &  0.6265 & 0.8703 & 0.7256 \\
\cmidrule(lr){2-8}
&mAP & \makecell{IN\_ODD + \\ EXTENDED\_ODD}& 61788 & 100504 &  0.6939 & 0.9669 & 0.8052 \\
\cmidrule(lr){2-8}
&e-mAP & \makecell{IN\_ODD + \\ EXTENDED\_ODD} &  61418 & 100504 &  \textbf{0.6976} & \textbf{0.9685} & \textbf{0.8072} \\

\bottomrule
\end{tabular}
\caption{Performance comparison between a yolov8 trained on \emph{In ODD} runways only, and one trained on \emph{In ODD} and \emph{Extended ODD} runways.}
\label{tab:in_extended_odd_comp}
\end{table*}

This section presents the experiments conducted to assess the factors influencing runway detection performance and generalization. Our evaluation is structured into two parts. First, we examine the impact of inclusion of the \emph{Extended ODD} runways in the training set on model performance and their ability to generalize across diverse conditions. Second, we analyze how the choice of source simulator in the training set affects the resulting models. 
All models are publicly available at~\cite{deelai-lard-v2-models}.  

\subsection{Influence of \emph{Extended ODD} runways}

As defined in Section \ref{sec:odd}, the intended function of the model is to detect runways for which the aircraft lies within the approach cone, referred to as \emph{In ODD} runways. A first naive baseline would therefore restrict the training set to these \emph{In ODD} runways as ground truth targets. However, learning to correctly reject \emph{Extended ODD} runways that lie very close to the ODD frontier can be challenging and may negatively affect the training. Moreover,  it is well established that ML models often generalize better when exposed to richer or more diverse training data. In this context, the \emph{Extended ODD} runways can be interpreted as a form of targeted data augmentation, providing additional geometric and visual variability around the operational boundary. For this reason, we compare two training strategies: one where only \emph{In ODD} runways are used as training target (model $M_{IN}$), and another where  \emph{In ODD} and \emph{Extended ODD} runways are treated as the same target class(model $M_{EXT}$). Performance and generalization are then assessed on the test set using the \emph{mAP} and \emph{e-mAP} (Section~\ref{sec:emap}). 
Experiments in this subsection use YoloV8 detectors trained with the Ultralytics framework~\cite{yolo11_ultralytics} under the same hyperparameters, the only difference being the composition of the training targets.


\begin{figure*}[h!bt]
    \centering
    \includegraphics[width=0.90\linewidth]{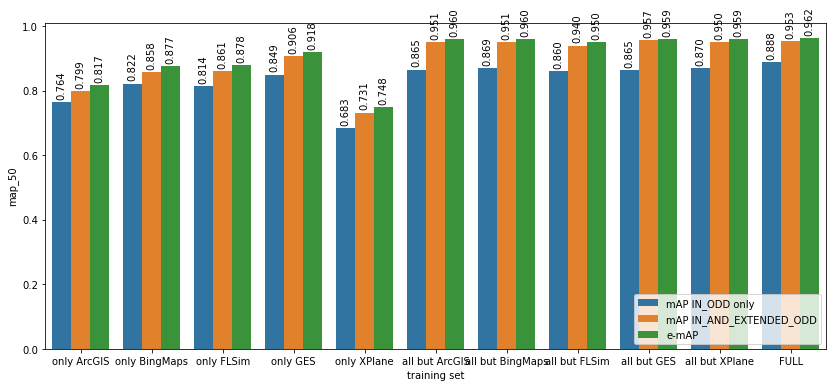}
    \caption{Comparison of \textit{mAP@50} performances for several training setup}
    \label{fig:all_training_sim}
\end{figure*}

Table~\ref{tab:in_extended_odd_comp} reports the results. On the strict \textit{In ODD} test subset, the model $M_{IN}$, trained on \emph{In ODD} runways,  reaches higher \textit{mAP} than $M_{EXT}$. This is expected, as $M_{IN}$ is trained to detect only \textit{In ODD} runways and tend to be more conservative, whereas $M_{EXT}$ detects many \emph{Extended ODD} runways that are ignored by the \textit{In ODD} metric and therefore counted as false positives, which in the end degrades the \emph{mAP} score. 
When considering the combined \textit{In ODD} and \textit{Extended ODD} test set, the situation reverses: the mAP decreases for $M_{IN}$ and increases for $M_{EXT}$, showing that including \textit{Extended ODD} examples in training leads to better overall performances once those near-boundary runways are considered valid targets.


The extended metric \textit{e-mAP} provides a more nuanced view of this trade-off. The column ``GT'' indicates the number of runways that were considered as ground truth in the metric computation. For both models, the values of the first two lines are indicated for reference, meaning we have 53610 runways \textit{In ODD}, and 8178 \textit{Extended ODD} runways (for a total of 61788 runways). However, for the \textit{e-mAP} rows, this value represents the maximum number of runways correctly detected, that resulted in the best value for the \textit{e-mAP} metric. Interestingly, $M_{IN}$ managed to correctly detect 4632 \textit{Extended ODD} runways (56,2\%), meaning it already generalizes fairly well to near-boundary configurations, despite never seeing them as targets. However, it is surpassed by $M_{EXT}$ which recovers the vast majority of these and almost doubles this number, detecting 7808 \textit{Extended ODD} runways (over 95,4\%). In other words, including \textit{Extended ODD} examples in training slightly degrades precision in the strict core of the ODD, but significantly improves generalization capability in a margin around the ODD boundary.


\subsection{Influence of simulator sources}
In this section we try to evaluate the influence of the simulator source on the training and generalization. For this we train eleven different yolov11 models under the same architecture and hyperparameters, varying only in the composition of training data. Five models are trained on images from a single source (ArcGIS, BingMaps, FLSim, GES and XPlane), five on all sources but one (leave-one-out configurations), and one on the union of all sources (FULL). Experiments in this subsection use YoloV11 detectors trained with the Ultralytics framework~\cite{yolo11_ultralytics} under the same hyperparameters, the only difference being the composition of the training targets.

Figure~\ref{fig:all_training_sim} summarizes the  \textit{mAP@50} performance of these models on the full test set. The FULL configuration consistently achieves the best average performance, which seem to indicate that combining multiple simulator and virtual globes during training is beneficial. Moreover, the fact that no leave-one-out configuration outperforms the FULL one confirms that each source effectively contributes in some way to the global performance. Among the single-source models, the X-Plane-only configuration clearly underperforms, while the GES-only achieves the highest mAP.

\begin{figure}[t!]
  \begin{subfigure}{\columnwidth}
    \centering
    \includegraphics[width=\linewidth]{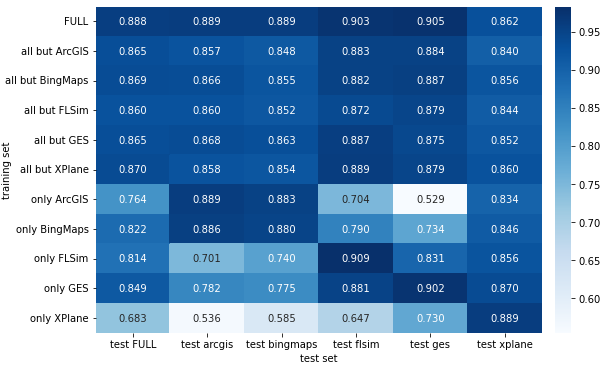}
    \caption{mAP on \emph{In ODD} runways}
  \label{subfig:tabular_crossbar_a}
    \vspace{0.7em}
  \end{subfigure}
  \begin{subfigure}{\columnwidth}
    \centering
    \includegraphics[width=\linewidth]{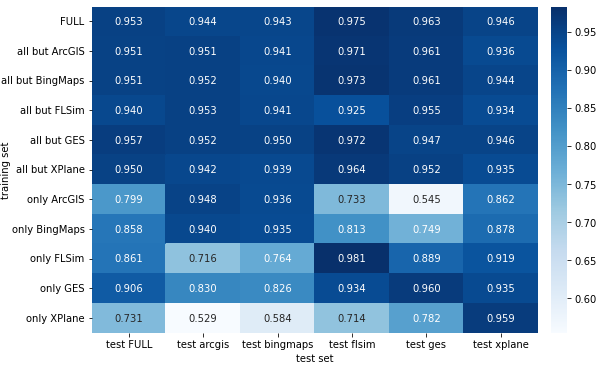}
    \caption{mAP on \emph{In ODD} and \emph{Extended ODD} runways}
  \label{subfig:tabular_crossbar_b}
    \vspace{0.7em}
  \end{subfigure}
  \begin{subfigure}{\columnwidth}
    \centering
    \includegraphics[width=\linewidth]{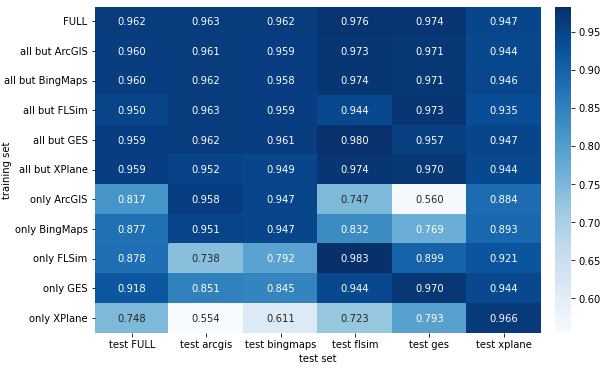}
    \caption{e-mAP}
  \label{subfig:tabular_crossbar_c}
    \vspace{0.7em}
  \end{subfigure}
  \caption{Crossbar \emph{mAP} performances of the eleven trained YOLOv11 models on each sub-test set.}
  \label{fig:tabular_crossbar}
\end{figure}

To better understand these behaviours, Figure~\ref{fig:tabular_crossbar} reports cross-source performance where each heat-map shows, for a given metric (\textit{mAP} on \textit{In ODD}, \textit{mAP} on \textit{In and Extended ODD}, and \textit{e-mAP}), how each model configuration performs when tested on images from each source. Three trends can be observed.

First, the X-Plane-only model performs well on X-Plane test images but poorly on all other sources. This suggests that its 3-dimensional models of runways and their textures lead to over-fitting to that particular domain, and that training only on such rendering generalize poorly to other virtual globes. Second, ArcGIS and Bing Maps models behave similarly and generalize reasonably well to each other, indicating proximity in their rendering and their domain. This was expected as these virtual globes have flat satellites texture mapped on the ground with altitude corrections. On the opposite, their results on Flight Simulator and GES reflect differences in textures, lighting and rendering between satellite imagery and 3D environments generated by simulators\footnote{Google Earth Studio provides 3D rendering of surrounding buildings, moving objects, as well as some nature elements on top of satellite imagery.}. Third, the GES-only model generalizes better than other single-source models to most other generators, which is consistent with the way it renders images, combining satellite imageries and 3D rendering.

Here again, the leave-one-out configurations lie between the single-source and FULL extremes, their performance being closer to that of FULL both in-domain and across domains. The relative ordering of models is also preserved when comparing  {mAP In ODD} and {e-mAP}, with the only case where a leave-one-out configuration outperforms the FULL is the \textit{all but GES} model on test images from Flight Simulator. However, this ordering frequently changes for \textit{mAP} computed on the combined \textit{In ODD} and \textit{Extended ODD} test set, as shown in Figure~\ref{subfig:tabular_crossbar_b}, which illustrates how unstable this metric can be in that specific context. This behaviour matches the issue identified in Section~\ref{sec:emap} and is precisely what our \textit{e-mAP} definition is designed to correct.



\section{Related works}\label{sec:related_works}

In this section, we first review pipelines for autonomous or assisted landing, then survey available runway datasets and simulators, and finally discuss studies that specifically rely on \lardun and its ODD.

\subsection{Vision-based landing pipelines}

Recent work on vision-based landing focuses on deep-learning pipelines that cover a large part of the approach-and-landing chain, from runway detection to pose estimation and guidance, and that are validated in realistic flight or simulator campaigns~\cite{balduzzi2021neural,ferreira2025vision}.
Other studies combine YOLO-like detectors, segmentation and geometric post-processing to derive relative pose and even close the loop with guidance and control in simulation~\cite{lin2025deep_auto_land,zhang2013ground}, while lightweight variants such as YOMO-Runwaynet target embedded fixed-wing platforms by combining YOLO-style heads with MobileNet-based backbones on dedicated runway datasets~\cite{dai2024yomo}. More recently, Ferreira~\emph{et~al.} presented an approach which uses a tracker and orientation estimator on runway imagery to generate guidance commands for fixed-wing UAVs~\cite{ferreira2025vision}.  Complementary work by Valen- -tin~\emph{et~al.} focuses on the downstream pose-estimation and assurance layers, first proposing probabilistic parameter estimators and calibration metrics for pose estimation from runway image features~\cite{valentin2024probabilistic}, and more recently introducing a real-time vision pipeline with calibrated predictive uncertainties and RAIM-based runtime monitoring for a computer vision landing system~\cite{valentin2025}. 
In parallel, Kouvaros~\emph{et~al.} apply formal verification techniques to semantic keypoint detection networks used for aircraft pose estimation, providing robustness guarantees for the perception stage of an autonomous landing system~\cite{kouvaros2023verification}. 
These works demonstrate that vision-only landing is technically feasible, but they typically rely on proprietary data or single-simulator sources, and they rarely expose the underlying dataset design or its Operational Design Domain (ODD) in detail.

\subsection{Runway datasets and simulators}

Public datasets for runway detection are still scarce. In 2023, the \lardun dataset was introduced as an open benchmark for front-view runway detection during approach and landing, combining a scenario generator based on Google Earth satellite imagery with a smaller set of real landing videos and, crucially, an explicit definition of the approach cone as an image-level ODD~\cite{ducoffe2023lard}. Around the same time, Chen~\emph{et~al.} proposed BARS, an airport runway segmentation benchmark built from X-Plane 11, with instance masks for runway and markings and a smoothing-oriented evaluation metric~~\cite{chen2023bars}. More recently, Wang~\emph{et~al.} introduced VALNet which included  a Runway Landing Dataset (RLD), also based mainly on X-Plane, targeting instance segmentation over full landing sequences and diverse weather conditions~\cite{wang2024valnet}. 
Beyond the runway domain, recent work on responsible dataset design stresses the importance of diversifying data sources and conditions to improve robustness and representation~\cite{data_creation_25}. 

Compared to these resources, \larddeux is designed explicitly around: (i) multi-runway airports selected among the world's busiest ones; (ii) interoperability between multiple virtual glo- -bes (Google Earth, Bing Maps, \arcgis) and flight simulators (\xplane, Flight Simulator); and (iii) an ODD-aware labelling scheme that distinguishes \emph{In ODD} and \emph{Extended ODD} runways in each image. While BARS and RLD primarily study segmentation and boundary quality within a single simulator, \larddeux focuses on object detection across heterogeneous generators, providing a complementary view on cross-source generalization.

\subsection{Work building on \lard}

Since its release, \lard has been used as a reference dataset for several methodological studies. YOLO-RWY~\cite{li2024yolo} introduces an anchor-free YOLO variant tailored to runway detection and reports significant gains on the \lard splits (synthetic, real-nominal, real-edge). Other papers use \lard to study robustness and security: a recent federated adversarial learning approach treats runway detection as a federated downstream task and evaluates adversarial resilience on LARD images~\cite{li2024federated}, while Zouzou~\emph{et~al.} propose a metric called  Conformal mean Average Precision (C-mAP) to equip YOLO-based detectors with statistical uncertainty guarantees on runway imagery~\cite{zouzou2025robust}. Finally, other lines of work leverage \lard for complementary goals, for instance BE-LIME~\cite{chen2025explainable}, which studies the explainability of runway detection models trained on it.

The ODD itself has also become an explicit object of study. Cappi~\emph{et~al.} describe a process to design and verify datasets with respect to a predefined ODD, translating system-level constraints into image-level Data Quality Requirements~\cite{cappi2024design}. The paper illustrates the whole process on \lard, for the ODD of a vision-based landing system. Our current paper can be seen as an extension of these efforts at the dataset level: we refine the original approach cone, introduce an \emph{Extended ODD} band around it, and propagate those labels consistently across our five data sources. On top of that, we define an extended detection metric (\textit{e-mAP}) that explicitly exploits the ODD structure and is orthogonal to the C-mAP from~\cite{zouzou2025robust}, so that uncertainty guarantees could be layered on top of our evaluation if needed.

\section{Conclusion}\label{sec:conclusion}

In this paper, we introduced \larddeux, an enhanced dataset and benchmarking framework for runway detection in vision-based autonomous landing systems. Compared to \lardun, the new version refines the Operational Design Domain by focusing on the approach segment, tightening and segmenting the approach cone, and extending it to multi-runway airports selected among the ones with the most traffic worldwide. On the data generation side, we proposed a calibrated workflow that uses a unified \texttt{.yaml} scenario description to produce accurately labelled images from multiple virtual globes (Google Earth, Bing Maps, \arcgis) and flight simulators (\xplane, Flight Simulator). This required a dedicated calibration procedure and an ODD-aware labelling scheme that separates \emph{In ODD} and \emph{Extended ODD} runways at the image level.

On the evaluation side, we defined an extended detection metric (e-mAP) that explicitly exploits this ODD structure by rewarding correct detections on both \emph{In ODD} and \emph{Extended ODD} runways. 
Using this framework, we reported experiments on two key factors: the impact of augmenting training data with new samples slightly beyond the ranges of the ODD, and the influence of 
simulator 
diversities.

The results show that training on both \emph{In ODD} and \emph{Extended ODD} examples improves robustness near the ODD boundary as captured by the \textit{e-mAP}, and that models trained on a mixture of all generators (FULL configuration) generalize substantially better across sources than single-simulator models, with X-Plane-only training proving particularly brittle. More generally, the proximity between generators also matters: virtual globes based purely on satellite textures (ArcGIS and Bing Maps) transfer well to one another but less to full 3D simulators, whereas hybrid tools which combine satellite imagery and 3D geometry (such as Google Earth Studio) tend to generalize better across both families.

Despite the recent growth of the literature, there is still no public dataset that combines multi-runway airports, multiple coordinated simulators and virtual globes, and an precise operational ODD with an associated detection benchmark; existing datasets typically rely on a single generator, and often consider single-runway settings with limited domain shift~\cite{ducoffe2023lard,chen2023bars,wang2024valnet}. Existing methodological work on top of \lard mostly treats the data and ODD as fixed. By releasing \larddeux and its calibrated multi-source generator, and by organising experiments around \textit{Extended ODD} training and simulator diversity, we aim to complement these works by providing an ODD-aware, multi-runway, multi-source benchmark on which models can be more meaningfully assessed, opening the way to follow-up studies on robustness, explainability and certification-oriented assurance methods. 

As a way forward, we plan to extend the present benchmark along two complementary axes. 
First, we are interested in a systematic comparison of different state-of-the-art object detection architectures (including transformer-based detectors, anchor-free models,  and different YOLO variants), as a thorough assessment of their respective strengths and weaknesses in the context of our well-defined ODD. Beyond global scores, the structure of this ODD, composed of three distance-based approach segments, will allow us to stratify evaluation by apparent runway size and scale, and thus to quantify, for each architecture, how performance evolves as the aircraft moves along the approach and as bounding boxes become larger and less ambiguous. 
Second, we intend to leverage the new capabilities offered by the support of modern flight simulators in \larddeux, to systematically vary clouds, weather conditions, illumination (including night-time scenarios) and seasonal appearance, and to generate dedicated subsets that extend the nominal ODD towards more complex and operationally relevant cases. Such situations had already emerged in \lardun, where they were handled through a manually curated "edge cases" subset of real images and videos. As demonstrated in~\cite{kraiem2025pose}, this approach will enable deeper analyses of model robustness under realistic conditions, and will support future work on gradually tightening assurance arguments as the ODD is extended beyond the current nominal approach segment.

\paragraph{Acknowledgement}
This work was carried out within the DEEL project,\footnote{\url{https://www.deel.ai/}} which is part of IRT Saint Exupéry and the ANITI AI cluster. The authors acknowledge the financial support from DEEL's Industrial and Academic Members and the France 2030 program – Grant agreements n°ANR-10-AIRT-01 and n°ANR-23-IACL-0002.
This work has also benefited from the PATO project funded by the French
government through the France Relance program, based on the funding from and by the European Union
through the NextGenerationEU program.

\bibliographystyle{plain}
\PHMbibliography{main}

\end{document}